\documentclass[letterpaper, 10 pt, conference]{ieeeconf}  %

\IEEEoverridecommandlockouts                              %

\usepackage{graphicx}
\usepackage{amsmath} %
\usepackage{subfig}
\usepackage{algpseudocode}
\usepackage{algorithm}
\usepackage{booktabs}
\usepackage{xspace}
\usepackage{multirow}
\usepackage{dblfloatfix} 
\usepackage{url}            %
\usepackage{hyperref}
\usepackage{amsfonts}

\title{\LARGE \bf
Efficiently Learning Small Policies for Locomotion and Manipulation
}

\author{Shashank Hegde and Gaurav S. Sukhatme\\
University of Southern California
\thanks{\footnotesize{\tt khegde|gaurav@usc.edu}}
\thanks{Sukhatme holds concurrent appointments as a Professor at USC and as an Amazon Scholar. This paper describes work performed at USC and is not associated with Amazon.}%
}%

\begin{document}

\maketitle

\thispagestyle{empty}
\pagestyle{empty}

\newcommand{\etc}{\emph{etc.}\xspace}
\newcommand{\ie}{\emph{i.e.,}\xspace}
\newcommand{\eg}{\emph{e.g.,}\xspace}
\newcommand{\etal}{\emph{et al.}\xspace}
\newcommand{\wrt}{with respect to\xspace}

\begin{abstract}

Neural control of memory-constrained, agile robots requires small, yet highly performant models. We leverage graph hyper networks to learn graph hyper policies trained with off-policy reinforcement learning resulting in networks that are two orders of magnitude smaller than commonly used networks yet encode policies comparable to those encoded by much larger networks trained on the same task. We show that our method can be appended to any off-policy reinforcement learning algorithm, without any change in hyperparameters, by showing results across locomotion and manipulation tasks. Further, we obtain an array of working policies, with differing numbers of parameters, allowing us to pick an optimal network for the memory constraints of a system. Training multiple policies with our method is as sample efficient as training a single policy. Finally, we provide a method to select the best architecture, given a constraint on the number of parameters. Project website:  
\href{https://sites.google.com/usc.edu/graphhyperpolicy}{https://sites.google.com/usc.edu/graphhyperpolicy}

\end{abstract}

\section{Introduction}
\label{sec:intro}

Deep reinforcement learning (RL) has produced exciting recent progress on diverse robotic tasks. State of the art results on applying deep RL to locomotion~\cite{haarnoja2019learning} or manipulation~\cite{gu2017deep} typically employ large networks ($\sim 256$ neurons, $\sim 3$ layers). In some applications \eg agile fast flight~\cite{song2021autonomous}, dynamic locomotion~\cite{2017-TOG-deepLoco} it is desirable to have highly performant, yet relatively small networks for two reasons: the onboard compute is limited (\eg to control a Crazyflie2.0 only 192Kb of RAM~\cite{batra2022decentralized} is available), and for agile tasks, the execution time for learned policies significantly affects overall performance~\cite{ibarz2021train}. Motivated by these constraints, we demonstrate a method to systematically generate small networks to solve robotic tasks using deep RL. Our method is both efficient (it does not increase sample complexity) and performant (within 80\%-90\% of peak performance while being 100x smaller than those previously reported). 

\begin{figure*}[h!]
    \centering

    \includegraphics[width=\linewidth]{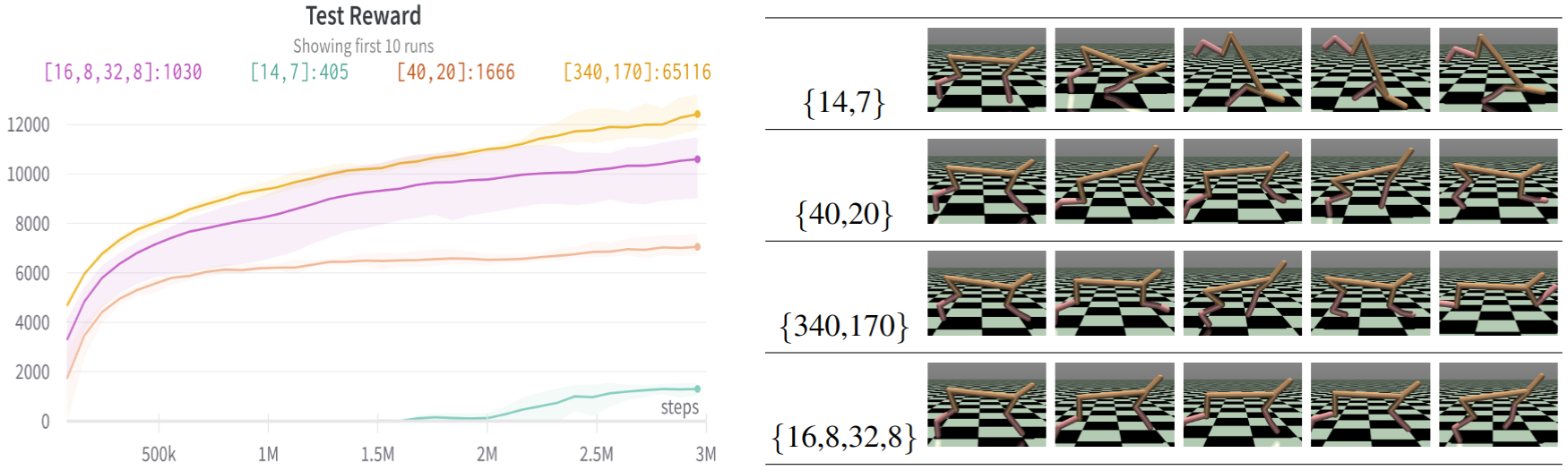}
    \captionof{figure}{Evaluating the effect of network architecture on the performance of control policies modeled by it. Trained and evaluated on the HalfCheetah task, a 2D robot with 6 control joints.}
    \label{fig:size_eval}
\end{figure*}

Finding small, yet performant networks for robotic tasks is non-trivial. As an illustration, consider Figure \ref{fig:size_eval} which compares performance across 4 networks of varying size, each trained using the Soft Actor Critic algorithm~\cite{haarnoja2018soft} on the HalfCheetah task~\cite{DBLP:journals/corr/BrockmanCPSSTZ16}, over 5 seeds.
The architectures of these networks are \{14,7\}, \{40,20\}, \{340,170\} and \{16,8,32,8\} respectively, where each number inside the parenthesis represents a hidden layer and the number of neurons within it. Performance is measured as the cumulative reward per rollout and is plotted across training till 3M data points are collected.  \{340,170\} (56116 parameters) achieves the highest performance. With size reduction \{40,20\} (1666 parameters) and \{14,7\} (405 parameters) there is degradation in performance. But interestingly, \{16,8,32,8\} (1030 parameters) performs significantly better than \{40,20\}, even though it has fewer parameters. This shows that arriving at a smaller performant network is not straightforward, and that performance depends not only the number of units, but the architecture of the network -- a point made in other contexts in neural network learning~\cite{JMLR:v20:18-598}. Further complicating matters, it is known that a reduction in performance with decreasing size may be partially attributed to lower exploration~\cite{pmlr-v100-mazoure20a}. In our work, we seek a method to achieve smaller networks without having to reduce exploration during training.

We learn a meta-policy, whose strategy and exploration is not limited by the representational capacity used to model it. We do this by leveraging graph hyper networks ~\cite{stanley2009hypercube,DBLP:conf/iclr/HaDL17,zhang2018graph,knyazev2021parameter} to learn architecture agnostic graph hyper policies, trained with standard off-policy RL. This results in thousands of functional approximators of the optimal policy, with a sample complexity comparable to learning just one, and models that estimate the optimal policy with fewer parameters. We demonstrate that for continuous control tasks, we can generate small policies that provide up to 90\% peak performance, without having to adjust any hyperparameters -- a practical time savings for someone searching for small networks. The networks we generate are spread across different sizes, allowing a choice of policy that conforms to a particular size constraint. This is done without consuming swaths of data, therefore being almost as sample complex as other state of the art RL algorithms that provide a single neural network policy. Further we show that graph hyper policies can be augmented with any off-policy RL algorithms, without changing any hyperparameters. 

We evaluate our trained policies on locomotion and manipulation tasks simulated in Mujoco~\cite{todorov2012mujoco}. 
\textbf{Our results show that smaller networks ($\sim$ 100x smaller than those traditionally used) trained with our method lead to 80-90\% performance of larger baseline networks.
Further, we show that a model's lossless memory dimension~\cite{DBLP:journals/corr/abs-1708-06019} is an accurate predictor of its eventual performance when estimated with our graph hyper policy.
}

\section{Related Work}
\label{sec:relwork}

\subsection{Hyper networks}
\begin{figure*}[ht]
    \centering
    \includegraphics[width=\textwidth]{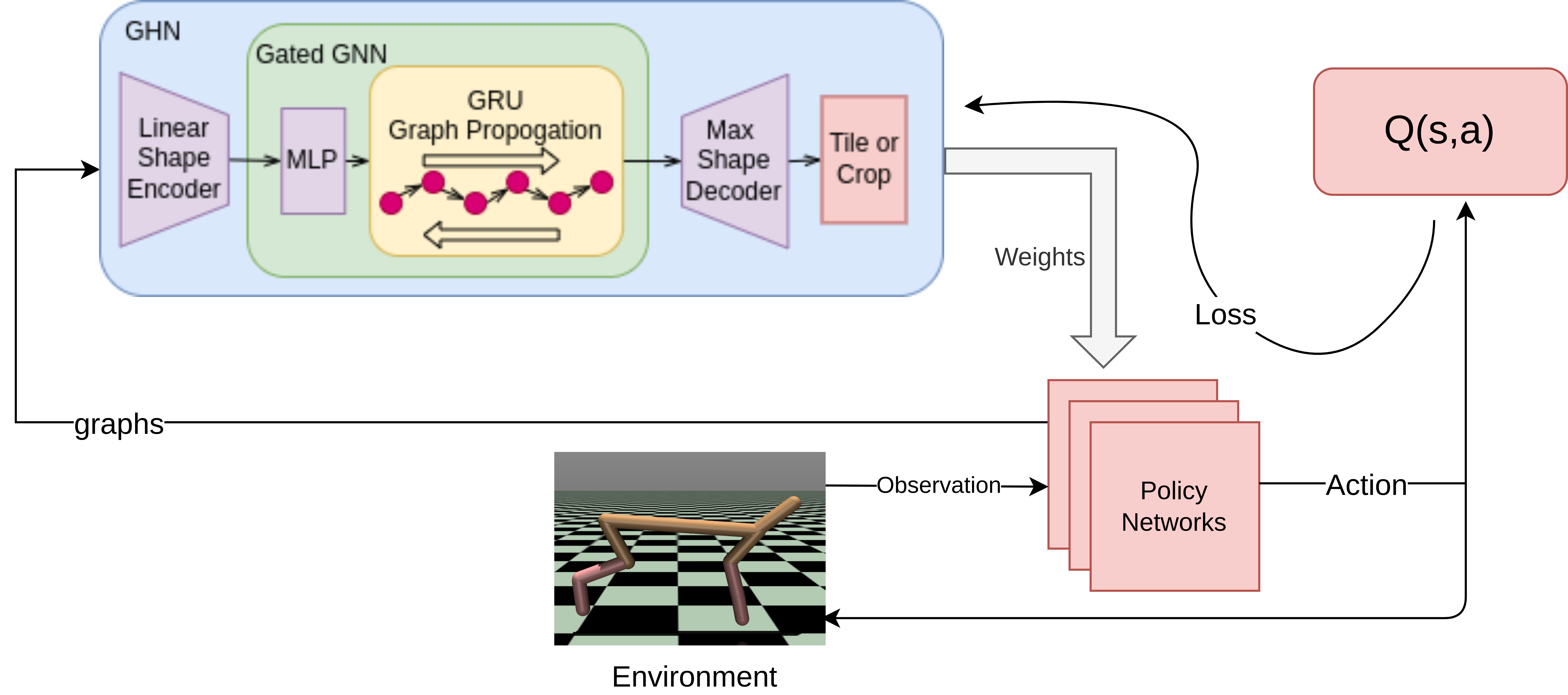}
    \caption{\textbf{Graph Hyper Policy.} Policy estimation using a graph hyper network (GHN) - based on the graphs of multiple policy network architectures, the GHN can estimate parameters for networks. The GHN can be optimized with an objective function of the policy network's action outputs.}
    \label{fig:hyper_net}
\end{figure*}

The main idea behind hyper networks~\cite{stanley2009hypercube} is to have a primary neural network (the hyper network) generate the weights for a secondary network. 
The input to the hyper network is the secondary network's architecture, its output is the set of weights for this architecture. It has been shown that it is possible to tune the weights of the hyper network with respect to the objective of a secondary network, using back propagation and stochastic gradient descent~\cite{DBLP:conf/iclr/HaDL17}. 
Graph hyper networks (GHN)~\cite{zhang2018graph} were shown to be effective at architecture search for image classification and generating weights for unseen model architectures (60\% accuracy on the CIFAR-10 dataset~\cite{knyazev2021parameter}). This was done by introducing a new dimension to the mini batch data, the meta batch, which samples a few architectures from a static dataset of 1M architectures. In our work, we use the secondary networks as policy networks to control robots.
Figure~\ref{fig:hyper_net} shows the underlying architecture of the GHN used here. 
In the graph representations of the secondary networks, each node is encoded to a higher dimension vector.
While the original paper uses a discrete embedded layer to do this, we use a simple linear encoder. 
This encoded graph is sent to a gated graph network (gated GNN).
The gated GNN consists of a multilayer perceptron that performs a nonlinear transformation of the encoded graph and a gated recurrent neural network (GRU). 
The GRU layer does a single forward and backward message pass between the transformed graph nodes.
From each transformed node, we decode a 2D matrix of (512,512). Based on the graph node's size (read layer size), we crop out a 2D matrix that will fit into that layer in the secondary network. 
This is repeated for all nodes in the graph.
Using propagation, the gradient from the loss function is used to train the entire GHN to correctly estimate the parameters for the secondary networks.
While this powerful tool has been previously used for supervised learning tasks, in this work we use it to solve robotic tasks with the help of reinforcement learning.

\subsection{Off-Policy Reinforcement Learning}
A key feature of off-policy RL algorithms~\cite{sutton2018reinforcement}, is to learn a policy from rollout data that was collected from different policies, while also collecting data for these different policies. This leads to sample efficient training.
The objective function is to maximize the cumulative future discounted rewards from the environment. 
The policy parameters $\theta$ are computed as follows.
\begin{equation}
    \theta^{*} = \arg \max_\theta \mathbb{E}\left[ \sum_{t=0}^\infty \gamma^t r_t\left(s_t, \pi_{\theta}(o_t)\right)\right]    
\end{equation}
where, $\pi_\theta$ is the policy, $\gamma$ is a discounting factor, and $r_t, s_t, o_t$ are the reward, state and observation at time t respectively. 
Since our method relies on changing architectures frequently while training, this family of RL algorithms is preferred over on policy and offline methods. 
We do not fix a particular off-policy algorithm in this work, although we do limit to actor critic methods for continuous control.
For locomotion tasks we use Soft Actor Critic (SAC)~\cite{haarnoja2018soft} as the baseline algorithm for training a policy.
For manipulation tasks, we employ a modified Deep Deterministic Policy Gradients (DDPG)~\cite{lillicrap2016continuous} algorithm to train policies.
Among the tasks we train on, manipulation tasks are significantly harder to solve, primarily because of sparsity of rewards in goal based tasks. For goal based tasks, one can employ hindsight experience replay (HER)~\cite{andrychowicz2017hindsight} to get high success rates. The idea is to modify the rollout data stored in the buffer and for a subset of trajectories, to set the achieved goal as the desired goal and recalculate the achieved reward.  This enables a DDPG agent to receive more positive reinforcement and learn faster.

\subsection{Lossless Memory Dimension for MLPs}
In this work, we only use multilayer perceptrons (MLP) to model policy networks.
To study the representational capacity of an MLP we use a metric called the Lossless Memory Dimension~\cite{DBLP:journals/corr/abs-1708-06019}, $D_{LM}$ defined as 
the maximum integer such that for any dataset with cardinality $n \le D_{LM}$ and points in random position, all possible labelings of this dataset can be represented with a function in the hypothesis space. This metric was originally meant to measure a MLP's capacity to classify, but we use it here to measure a model's capacity to regress to optimal action distribution parameters. 
Algorithm~\ref{alg:cap} shows how we calculate this metric (a minor modification to~\cite{DBLP:journals/corr/abs-1708-06019} by adjusting for multivariate regression (line 5)). 
The intuition behind this addition is that classification is similar to regression without the final softmax (or sigmoid) layer. 

\begin{algorithm} \caption{Calculate $D_{LM}$ of an MLP}\label{alg:cap}
\begin{algorithmic}[1]
\Require  
\Statex {$Inp$ - Input dimensions of MLP}
\Statex {$Out$ - Output dimensions of MLP}
\Statex {$a_j = \{l_i\}_1^k$ - MLP Architecture}

\State {Initilize $D_{LM} = (Inp + 2) * l_0$;} 
\For {$j \leftarrow 2, k$}
\State {$D_{LM} \leftarrow D_{LM} + minimum(\{ l_i \}_1^j)$}
\EndFor
\State {$D_{LM} \leftarrow D_{LM} + minimum(\{ l_i \}_1^k , Out)$}
\end{algorithmic}
\end{algorithm}

\section{Method}
\label{sec:method}

\subsection{Graph Hyper Policy}
A common practice to formulate a stochastic policy in RL is to model it as neural network that accepts the agent (robot's) observation, and output a mean and variance vector, to represent an action distribution. 
We use a GHN (Figure~\ref{fig:hyper_net}) to estimate the parameters needed to model a policy network, given an architecture graph. 
This policy network maps observation to action.
Following~\cite{knyazev2021parameter}, the graph hyper policy is capable of accepting a meta batch of architecture graphs, and estimating multiple parametric policies at the same time.

\subsection{Architecture Sampling}
We restrict ourselves to feedforward sequential MLPs with upto 4 layers, to model a policy. 
For each layer, the number of neurons can take values from the set $\mathcal L $.
For our experiments, $ \mathcal L = \{4,8,16,32,64,128,256,512\}$. 
The set of all possible MLPs is denoted by $\mathcal A$ and their corresponding graphs by  $\mathcal G$,
\begin{equation}\label{eq:G}
    \mathcal A = \{a_j\}, \mathcal G = \{g_j\},   1\leq j \leq 4680
\end{equation}
 
We denote each unique architecture as $a_i$ and its computational graph as $g_j$.

\begin{equation}\label{eq:g}
  \centering
 \left.
    \begin{array}{ll}
        a_j = \{ l_i \}_{i=1}^k\\
        g_j = [0, \{ l_i, l_i \}_{i=1}^k , 2a]\\
    \end{array}
\right \} {1 \leq k \leq 4}
\end{equation}

where, node $ l_i \in \mathcal L $. 
In $g_i$ the first value is 0, which denotes the input node and the output node is represented by $2a$, where $a$ is the action space dimension. 
The multiplication factor 2 enables our MLPs to represent a stochastic policy with both action mean and variance vectors. 
Each $l_i$ is repeated twice in $g_i$ as this represents the weights and biases nodes. 
Note $| \mathcal G|$ is 4680 -- this is the total number of unique architectures possible with upto 4 layers with each layer in $\mathcal L$\\

\subsection{Training Procedure}
In most state of the art deep RL implementations, rollout data are generated from a vector of parallel environment instances. 
This speeds up training greatly. 
In our method, we set this number of parallel instances as the meta batch size. 
Thus, during rollout, each architecture samples data from its own corresponding environment instance.
We store the transition data from all environments to a common buffer. 
For locomotion tasks, during rollout, every few steps, we sample a new meta batch of architectures, $g_i \sim  Uniform(G)$. 
We take care not to lower this number significantly as this can make the rollout data very noisy and destabilize training. 
For manipulation, we sample new architectures only at the beginning of each rollout.

We use a simple critic and critic target network. This network models the Q value function. We sample a batch of data from the replay buffer and perform a off-policy Bellman update on the critic network. For locomotion tasks, we add an entropy regularizer. From the minibatch, we split the data among the sampled architectures, obtain corresponding actions and update the GHP by maximizing their Q values.

\begin{equation}
    \arg\max_{\theta} E_{s}[Q(s, a)]  \quad \quad a \sim \phi_i (\cdot | s) \quad \quad \phi_i = h_{\theta}(g_i)
\end{equation}
where, $h_{\theta}$ denotes the GHP. 
For a sampled graph $g_i$ from the metabatch, we estimate a parametric state conditioned action distribution $\phi_i$ modeled by the corresponding MLP, using $h_{\theta}$. 
We then sample an action $a$ from this distribution. 
We optimize over the Q value of this action across sampled actions from the minibatch, by adjusting the GHP parameters $\theta$. 
This optimization, like other actor critic reinforcement learning algorithms, is done using stochastic gradient descent. 
Originally, \cite{knyazev2021parameter} suggested using a learning rate scheduler during SGD for the hyper network for image classification, but we see that this is not needed for GHP training.

\subsection{Parameter Constrained Architecture Prediction}
We are motivated to find smaller control policies for robotic tasks. This is done after we place a constraint on the maximum number of parameters used to model the policy.  After training a graph hyper policy, one way to find the best architecture  is to perform an exhaustive search across all architectures with fewer parameters than the maximum number allowed. This entails estimating parameters for each candidate architecture, evaluating it by performing rollouts, and calculating its cumulative reward. This process is computationally intensive. It is therefore reasonable to ask if the best architecture (or an architecture with performance close to the best) can be found efficiently. We evaluate two options. The first is to simply find the architecture with the largest number of parameters that satisfies the size constraint and estimate its parameters with the hyper network. The second is to use Equation~\ref{eq:best_mlp} which utilizes the lossless memory dimension $D_{LM}$ (Algorithm~\ref{alg:cap}, Section~\ref{sec:relwork}) to predict the best architecture. 
\begin{equation}\label{eq:best_mlp}
    a_{opt} = argmax_{a_i}(D_{LM}(Inp, Out, a_i))
\end{equation}

We show that for a given parameter count constraint, among the architectures trained by the GHP, those that maximize the lossless memory dimension perform better than the largest architecture satisfying the parameter constraint. Empirical support for this claim is presented in the following section.
\newcommand\plotimagewidth{60.0mm}

\begin{figure*}[h]
\centering
\setlength{\tabcolsep}{0.0mm}
\begin{tabular}{c c c}
{\includegraphics[width=\plotimagewidth]{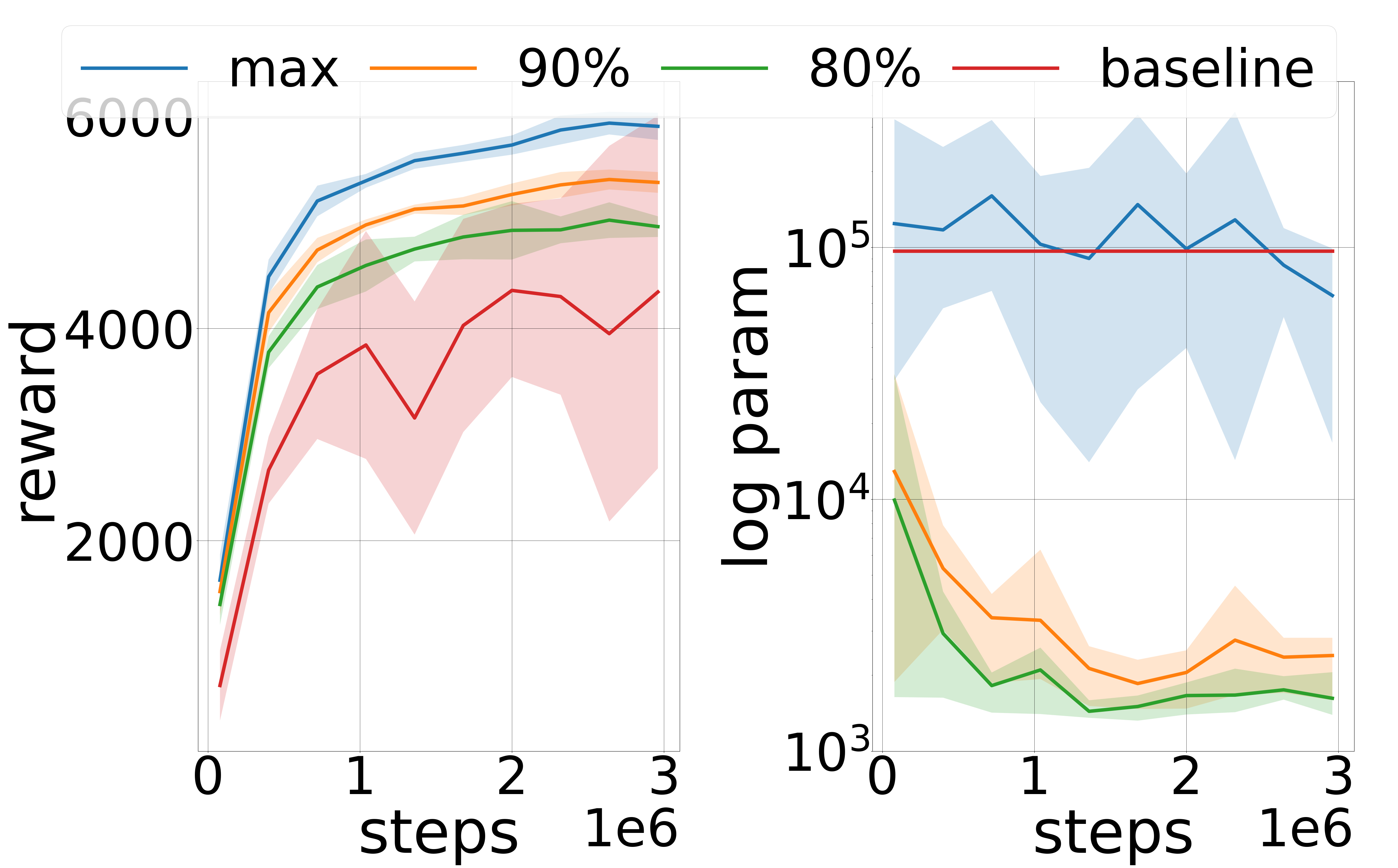}} &  {\includegraphics[width=\plotimagewidth]{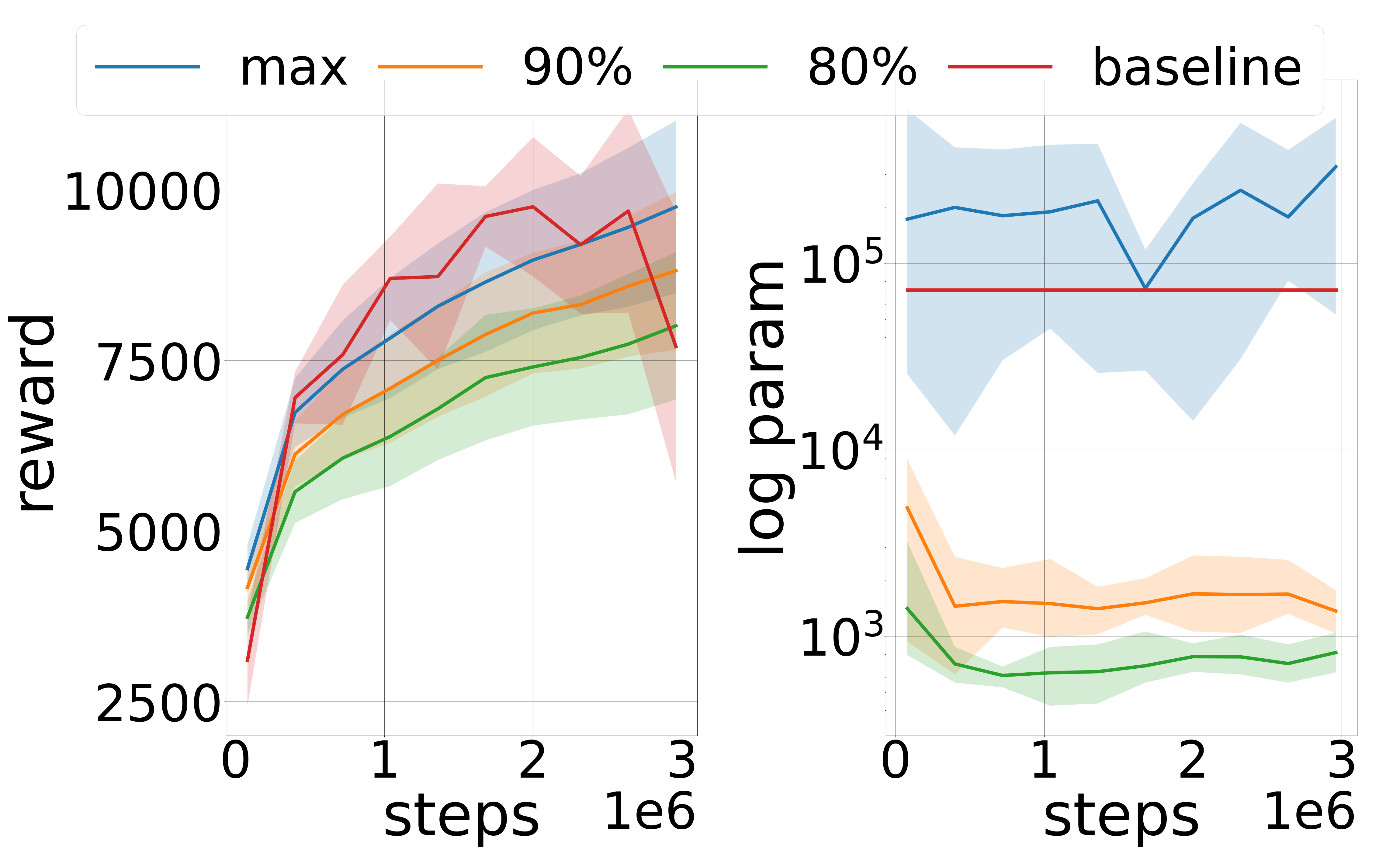}}  & {\includegraphics[width=\plotimagewidth]{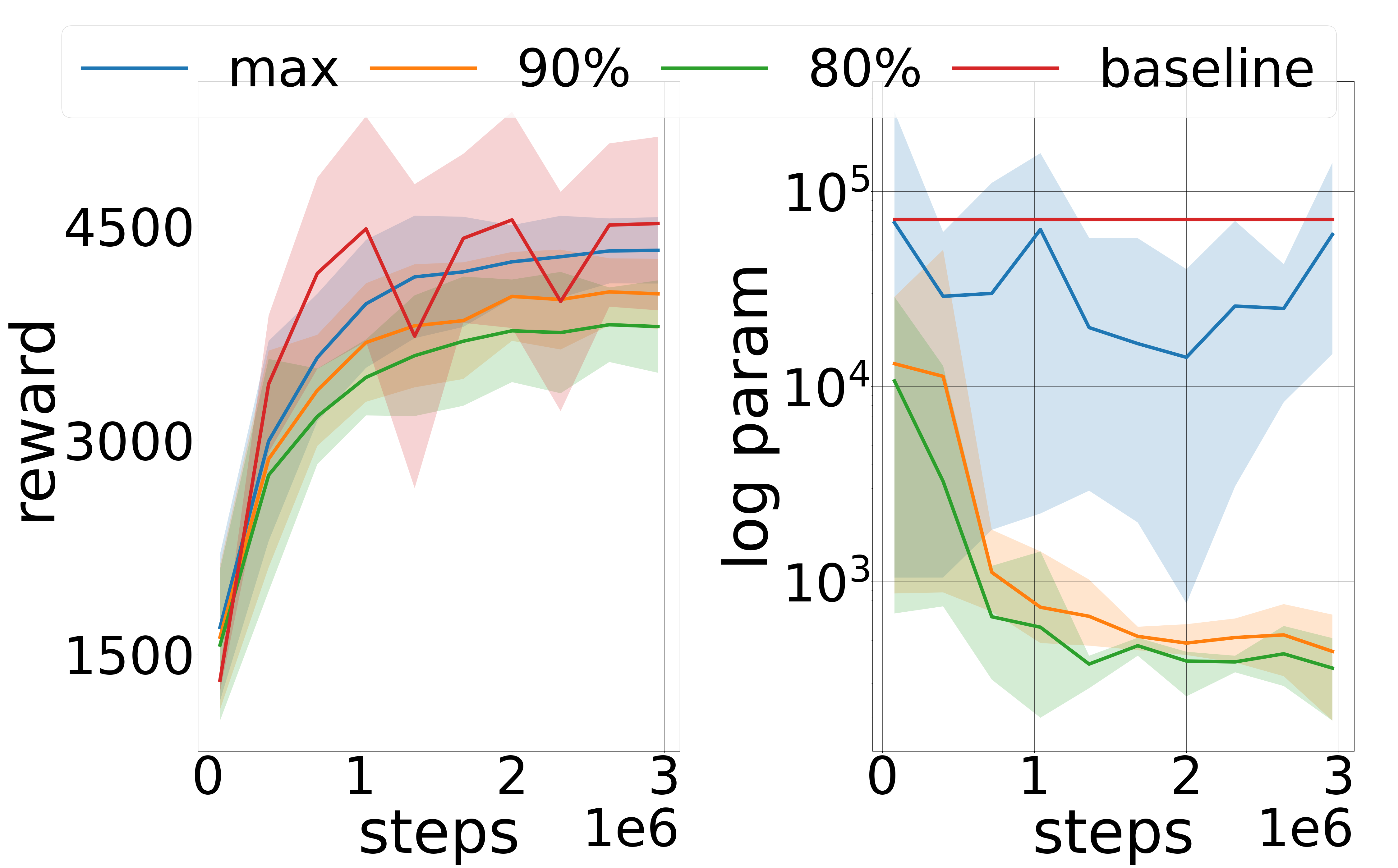}} \\

{(a) Ant-v2} & {(b) HalfCheetah-v2} & {(c) Walker2d-v2} \\[6pt]

 {\includegraphics[width=\plotimagewidth]{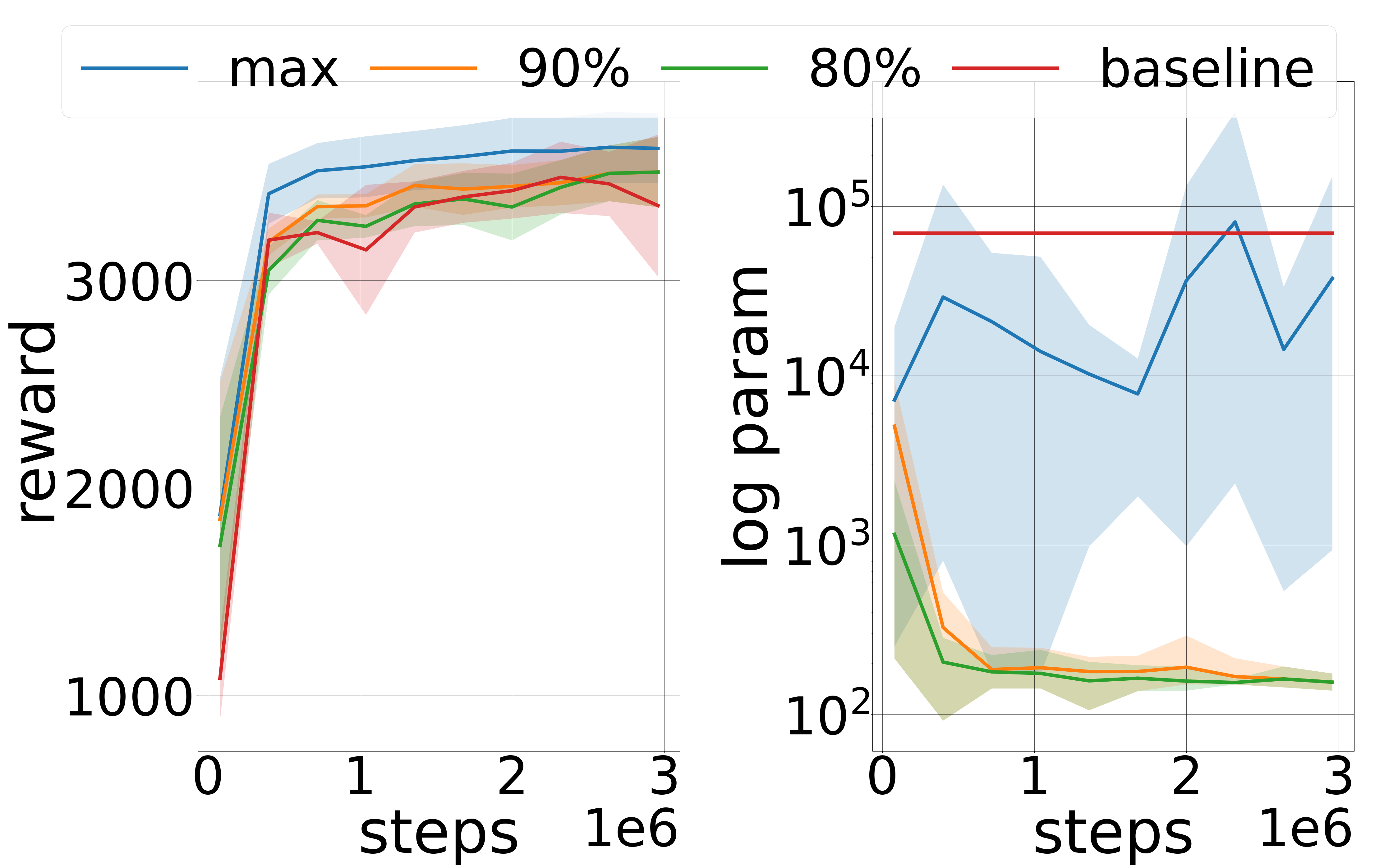}} & {\includegraphics[width=\plotimagewidth]{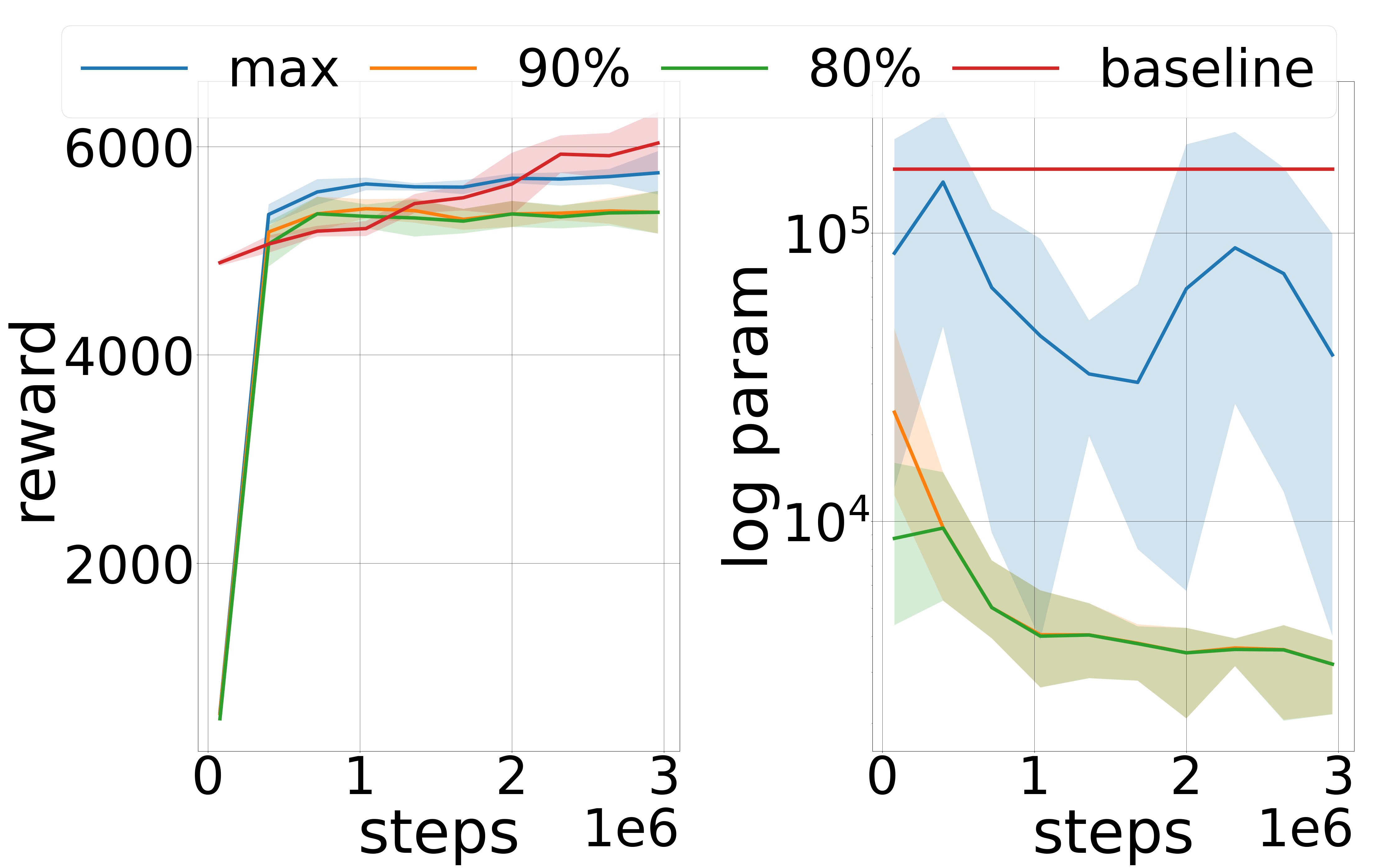}} & {\includegraphics[width=\plotimagewidth]{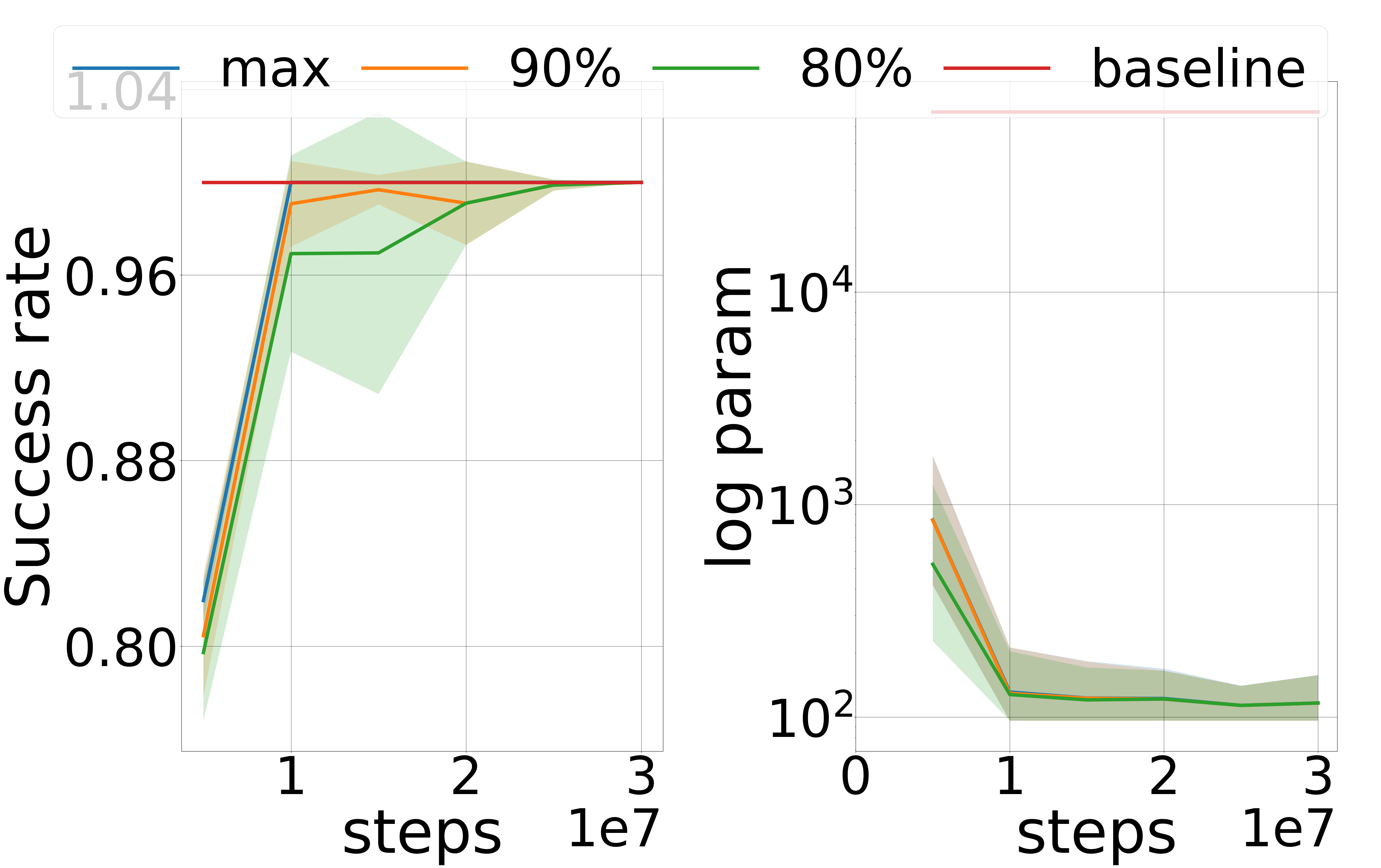}} \\
 
{(d) Hopper-v2} & {(e) Humanoid-v2} & {(f) FetchReach-v1} \\[6pt]

 {\includegraphics[width=\plotimagewidth]{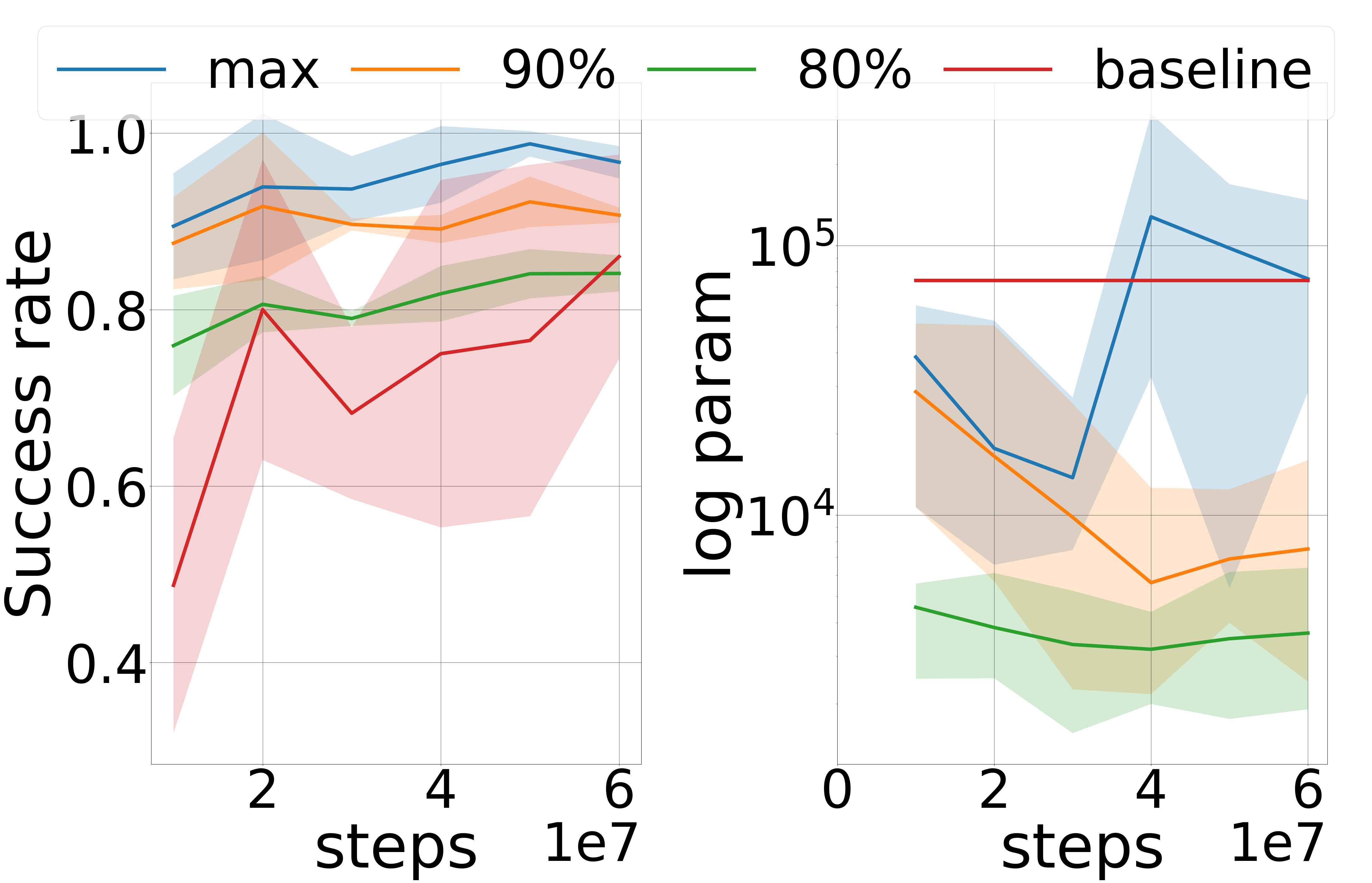}} & {\includegraphics[width=\plotimagewidth]{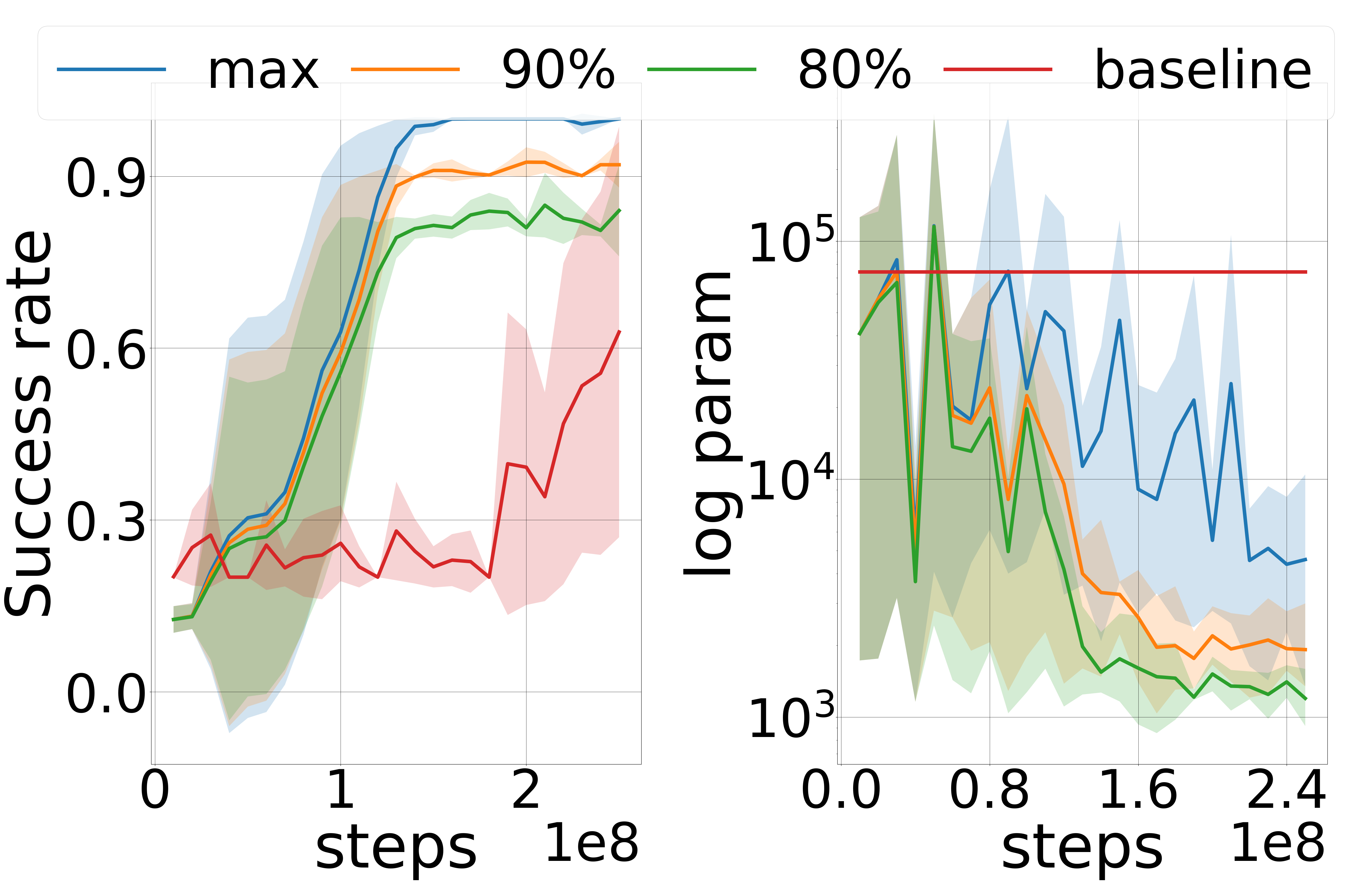}} & {\includegraphics[width=\plotimagewidth]{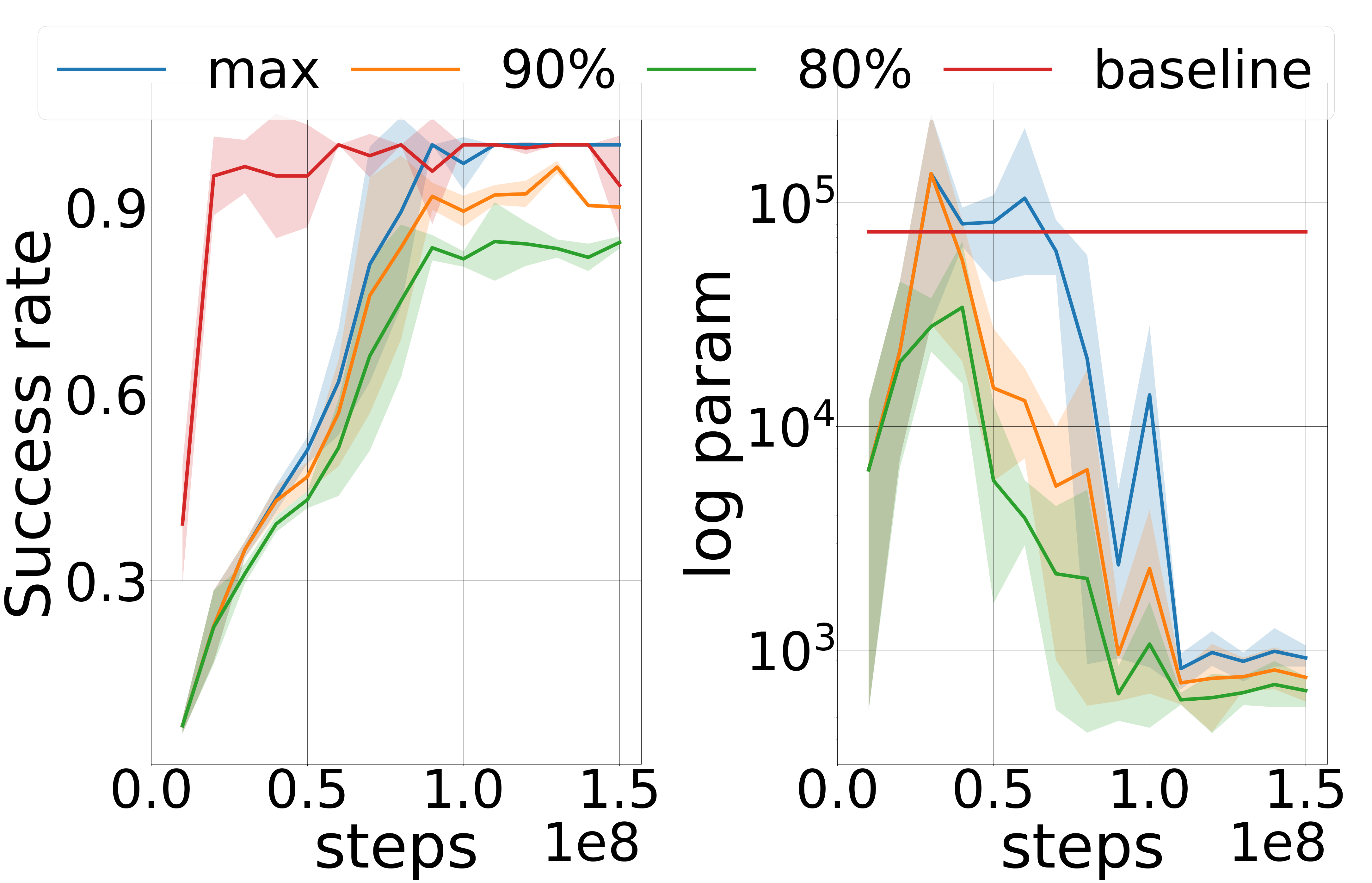}} \\

{(g) FetchSlide-v2} & {(h) FetchPush-v1} & {(h) FetchPickAndPlace-v1} \\[6pt]

\end{tabular}

\caption{\textbf{Learning smaller networks.} All architectures evaluated as training progresses. For each pair, \textbf{left}: (max performance, 90\% of max performance, 80\% of max performance, baseline performance) vs training samples collected; \textbf{right}: the minimum number of parameters needed to achieve these levels of performance vs training samples collected.}
\label{fig:percent_plots}
\end{figure*}

\algnewcommand\algorithmicforeach{\textbf{for each}}
\algdef{S}[FOR]{ForEach}[1]{\algorithmicforeach\ #1\ \algorithmicdo}

\section{Results and Discussion}
\label{sec:expts}

\begin{table*}[]
        \centering
            \setlength{\tabcolsep}{1.45mm} %

        \begin{tabular}{l | l || c | c || c | c | c }
            \multicolumn{2}{c}{} & \multicolumn{4}{c}{Method} \\
            \cmidrule{3-7}
            \multicolumn{2}{c}{Task} & Baseline & GHP (ours), max  & Behavior cloning &  Off-Policy RL & GHP (ours), 90\%\\
            \hline \hline
            Ant & Architecture & \centering [256, 256] & \centering [256, 512, 64, 8] & [16,32,32,4] & [16,32,32,4] & [16,32,32,4] \\
             & Parameters & 96520 & 193680 & 3564 & 3564 & 3564 \\\cmidrule{3-7}
             & Reward & $4343.35 \pm 1662.70$ & \textbf{5746.23} $\pm$ \textbf{102.95} & $-2984.74 \pm 1330.11$ & $5180.39 \pm 397.19$ & \textbf{5226.22} $\pm$ \textbf{433.03} \\
            \hline

            HalfCheetah & Architecture & [256, 256] & [128, 512, 512, 256] & [64,4,4,32] & [64,4,4,32] & [64,4,4,32]  \\
             & Parameters & 71942 & 463878 & 1790 & 1790 & 1790 \\\cmidrule{3-7}
            & Reward & $7703.76 \pm 1980.54$ & \textbf{9438.79} $\pm$ \textbf{1324.98} & $-1582.32 \pm 2156.98$ & $6879.44 \pm 1711.91$ & \textbf{8379.85} $\pm$ \textbf{1101.69} \\
            \hline

            Walker2d & Architecture & [256, 256] & [32, 256, 4, 64] & [16,16,4] & [16,16,4] & [16,16,4] \\
             & Parameters & 71942 & 10762 & 658 & 658 & 658 \\\cmidrule{3-7}
            & Reward & \textbf{4517.07} $\pm$ \textbf{608.14}  & 4175.81 $\pm$ 233.13 & $814.05 \pm 110.55$  & $3655.73 \pm 536.59$ & \textbf{3825.65} $\pm$ \textbf{401.81} \\
            \hline

            Hopper & Architecture & [256, 256] & [32, 512, 4, 32] & [8,4,4,4] & [8,4,4,4] & [8,4,4,4] \\
             & Parameters & 69635 & 19591 & 187 & 187 & 187 \\\cmidrule{3-7}
            & Reward & 3359.96 $\pm$ 342.01 & \textbf{3607.88} $\pm$ \textbf{181.18} & $ 1174.50 \pm 187.42$ & \textbf{3467.56} $\pm$ \textbf{179.06} & 3335.17 $\pm$ 313.61 \\
            \hline

            Humanoid & Architecture & [256, 256] & [16, 8, 32, 16] & [8,32,8] & [8,32,8] & [8,32,8]  \\
             & Parameters & 166673 & 7273 & 3721 & 3721 & 3721 \\\cmidrule{3-7}
            & Reward & \textbf{6034.41} $\pm$ \textbf{300.53} & 5472.11 $\pm$ 161.74 & $4888.40 \pm 20.54$ & \textbf{5418.53} $\pm$ \textbf{149.53} & 5371.20 $\pm$ 277.34 \\
            \hline \hline

            FetchReach & Architecture & [256, 256] & [4, 8] & [4, 8] & [4, 8] & [4, 8]  \\
             & Parameters & 70404 & 132 & 132 & 132 & 132 \\\cmidrule{3-7}
            & Success Rate & \textbf{1.0} $\pm$ \textbf{0.0} & \textbf{1.0} $\pm$ \textbf{0.0} & $0.84 \pm 0.23$ & \textbf{1.0} $\pm$ \textbf{0.0} & \textbf{1.0} $\pm$ \textbf{0} \\
            \hline

            FetchSlide & Architecture & [256, 256] & [256, 128, 128, 16] & [64,64,32,16] & [64,64,32,16] & [64,64,32,16] \\
             & Parameters & 74244 & 58964 & 8692 & 8692 & 8692 \\\cmidrule{3-7}
            & Success Rate  & \textbf{0.86} $\pm$ \textbf{0.110} & 0.82 $\pm$ 0.10 & 0.12 $\pm$ 0.24 & \textbf{0.76} $\pm$ \textbf{0.11} & \textbf{0.74} $\pm$ \textbf{0.10}  \\
            \hline

            FetchPush & Architecture & [256, 256] & [16,256,16,32] & [32,32,8,16] & [32,32,8,16] & [32,32,8,16] \\
             & Parameters & 74244 & 9604 & 2460 & 2460 & 2460 \\\cmidrule{3-7}
            & Success Rate  & 0.62 $\pm$ 0.35 & \textbf{0.89} $\pm$ \textbf{0.10} & 0.2 $\pm$ 0.01 & 0.31 $\pm$ 0.427 & \textbf{0.81} $\pm$ \textbf{0.06}  \\
            \hline

            FetchPickAndPlace & Architecture & [256, 256] & [16, 16, 8] & [16, 16, 4] & [16, 16, 4] & [16, 16, 4] \\
             & Parameters & 74244 & 908 & 824 & 824 & 824 \\\cmidrule{3-7}
            & Success Rate  & $0.93 \pm 0.08$ & \textbf{1.0} $\pm$ \textbf{0.0} & $0.04 \pm 0.08$  & \textbf{0.98} $\pm$ \textbf{0.03} & 0.91 $\pm$ 0.17 \\
            \hline \hline

           \end{tabular}
       \caption{\textbf{Comparison of performance.} Baseline corresponds to the performance of a standard [256,256] network trained using off-policy learning. GHP max corresponds to the best performing network estimated by the GHP. Our best performing networks are on par with the baseline. We identify smaller architectures that achieve at least 90\% performance. We compare models with these architectures trained with behavior cloning, off-policy RL, and that estimated by our GHP. }
       \label{tab:perf}
\end{table*}

We evaluate our method on locomotion and manipulation tasks simulated in Mujoco.
We experiment with the following locomotion tasks: HalfCheetah (described earlier); Ant (a 4 legged robot with observation dimension 111 and action dimension 8); Walker2D (a 2D biped with observation dimension 17 and action dimension 6); Hopper (a 2D robot with 3 joints, with observation dimension 11 and action dimension 3); and Humanoid (a walker, with observation dimension 376 and action dimension 17). 
For these tasks, forward movement is rewarded while energy consumption and large orientation changes are penalized. For locomotion tasks, we train the GHP for 3M steps over 8 parallel environment instances, using Soft Actor critic. 

For the manipulation tasks, we use the Fetch environments~\cite{DBLP:journals/corr/abs-1802-09464}. 
These environments simulate goal based tasks on the Fetch robotic arm. 
For our evaluation, we use Reach (a simple task that rewards moving the end effector to a goal location, observation dimension 10); Slide (the arm has to move a puck on a slippery table without using the gripper, observation dimension 25); Push (the arm has to move an object on a 2D surface without picking it) and PickAndPlace (the arm has to move an object in 3D space from a starting point to a goal location, observation dimension 25). 
For all these tasks, the action dimension is 4 representing the displacement of the end effector in 3D and gripper state. 
We also append the desired goal to the observation while querying the policy network for actions.
For these tasks, we  provide sparse reward (only when the goal is achieved) and we track the success rate for achieving the desired goal across multiple evaluations.
For the FetchReach task we train for 30M steps; FetchSlide for 60M steps; FetchPush for 250M steps; and FetchPickAndPlace for 150M steps. We run 8 environment instances in parallel, and train using DDPG with HER. For each task, our baseline is a single $[256, 256]$ MLP policy network, trained with standard off-policy learning algorithms. For all our runs we re-implement these algorithms with our training method. The only difference between GHP training and the baseline is the choice of policy network, every other hyperparameter remains untouched. All results are averaged over 5 seeds.

\subsection{Finding High Performing Small MLPs}
At every few checkpoints, we analyse GHP performance. We evaluate it on all 4680 architectures with 5 rollouts and find the average total reward for each architecture. 
We find the smallest architecture that provides the (a) maximum reward, (b) at least 90\% of the maximum reward and (3) at least 80\% of the maximum reward. 
We store these architectures, their corresponding number of parameters and their average reward on rollout. We term these architectures {max, 90\% and 80\%} architectures. 
Figure~\ref{fig:percent_plots} shows this analysis for all tasks. 
In each image pair per task, the left image shows the progress of the average cumulative reward of the {max, 90\% and 80\%} architectures at each checkpoint. 
The image on the right corresponds to the log of the number of parameters of the {max, 90\% and 80\%} architectures.
For each task, the number of samples to train 4680 networks with the GHP is comparable to that needed by the baseline to train one network. 
This shows that there is positive transfer of knowledge between each architecture in the training set. 

For the Ant task, even the 80\% architectures seem to assimilate higher average cumulative reward than the baseline. 
For the Walker2D and Hopper tasks, our architectures perform similar to the baseline. 
On the HalfCheetah task, none of our architectures seem to perform as well as the baseline suggesting that generalization between architectures requires more samples and further training for this task.
Finally, on the Humanoid control task, we converge to a working policy faster than the baseline. We conjecture that this is because sampling from different architectures helped with exploration during early stages of training. Eventually the baseline achieves better rewards.

With manipulation, we see similar results. The FetchReach task is simple -- the baseline quickly achieves a success rate of $1.0$. 
Our method is not as sample efficient as the baseline, but achieves a success rate of $1.0$ for almost all 4680 architectures. The FetchSlide task is significantly harder to solve. Here our method performs better than the baseline and provides smaller networks that perform as well as the baseline with better sample efficiency. A similar observation can be seen with FetchPush.
For FetchPickAndPlace, our method requires more samples than the baseline. But we eventually do arrive at significantly smaller policies with success rate of $1.0$. For each task, as training progresses, 90\% and 80\% architectures are orders of magnitude smaller than max architectures. At the end of training, these architectures are saved as highly performant small MLPs.

\begin{figure}
\centering
\begin{tabular}{p{4cm} p{4cm}}
  \includegraphics[width=40mm]{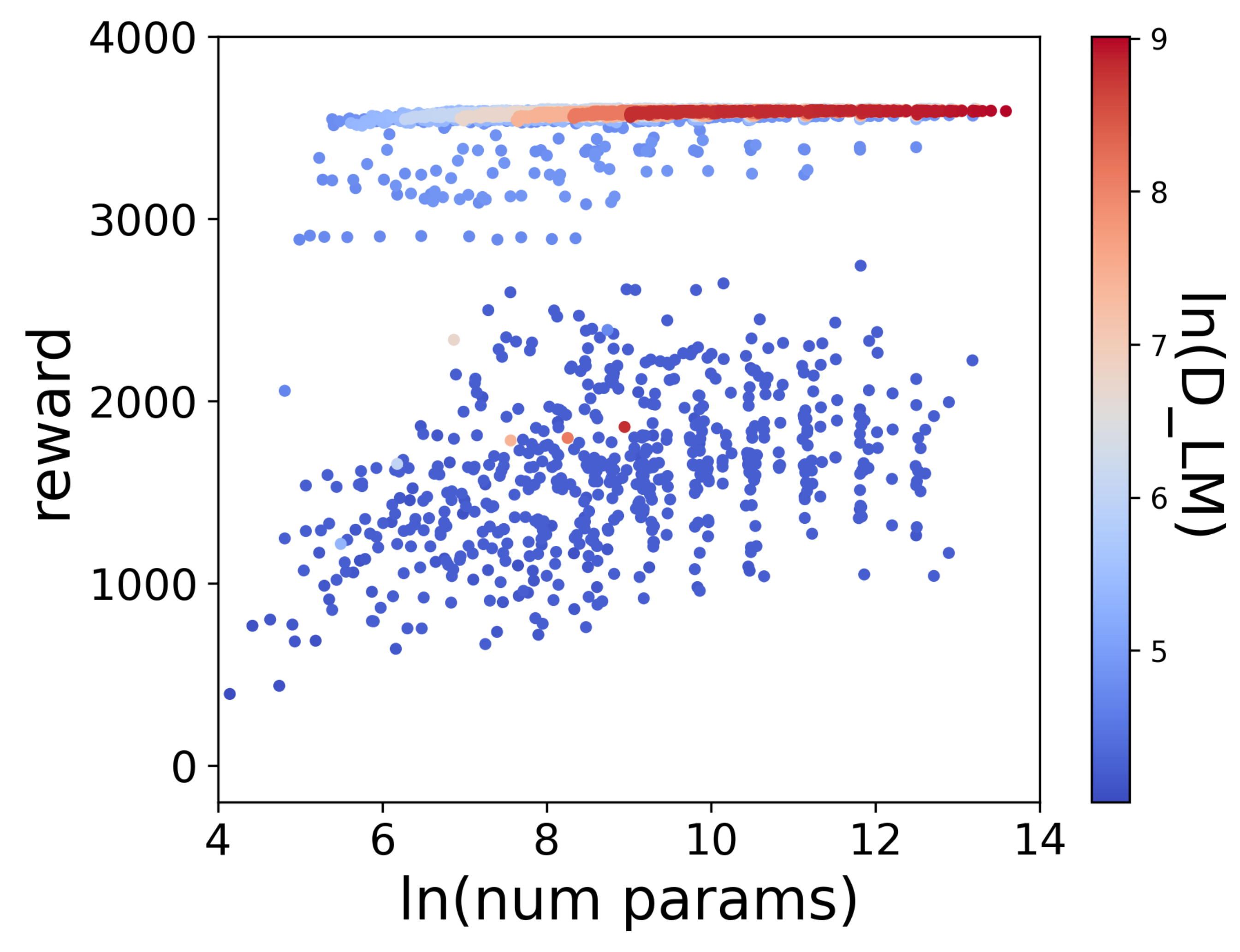} & \includegraphics[width=40mm]{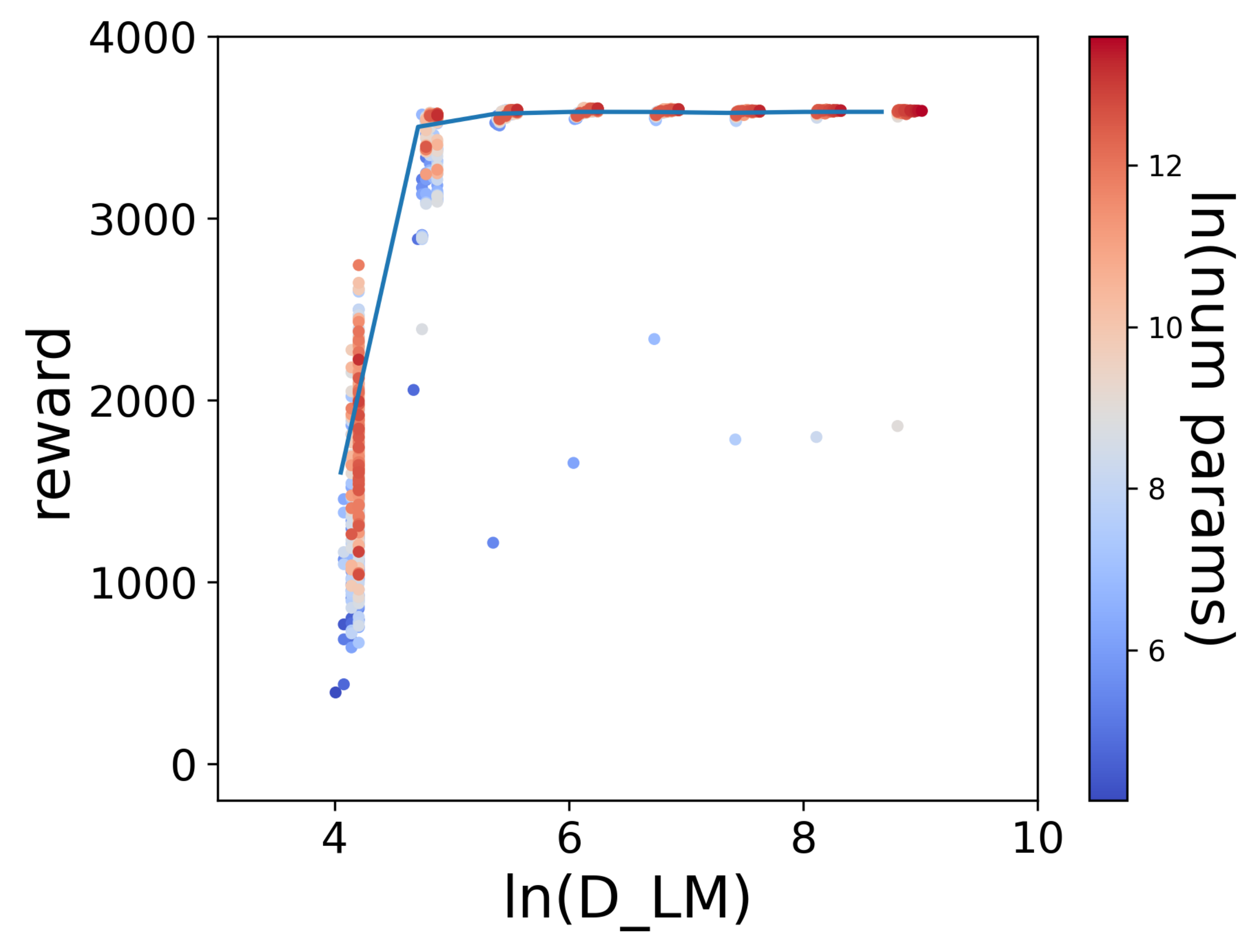}  \\
  \multicolumn{2}{c}{ (a) Hopper-v2}\\[6pt]
  \includegraphics[width=40mm]{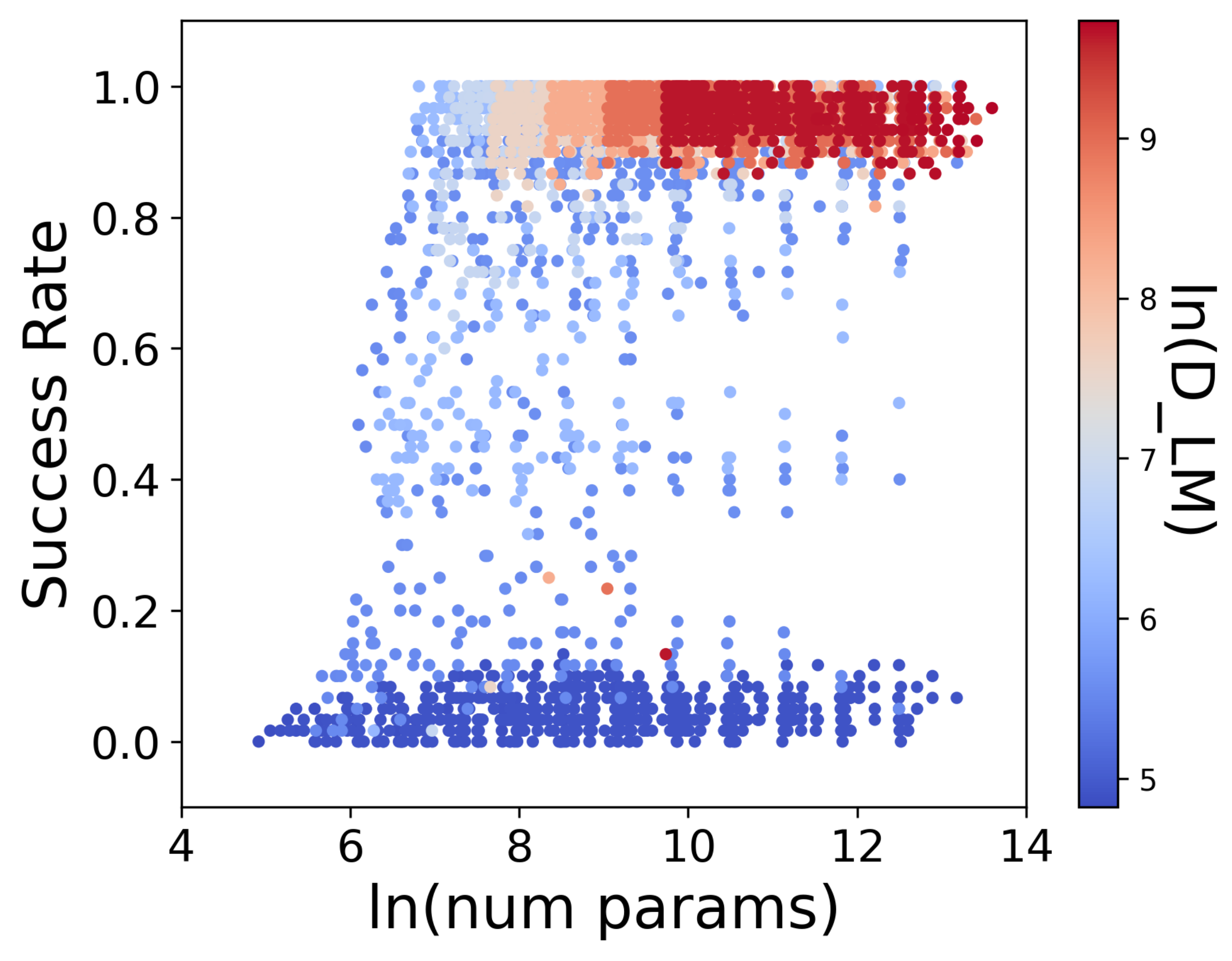} & \includegraphics[width=40mm]{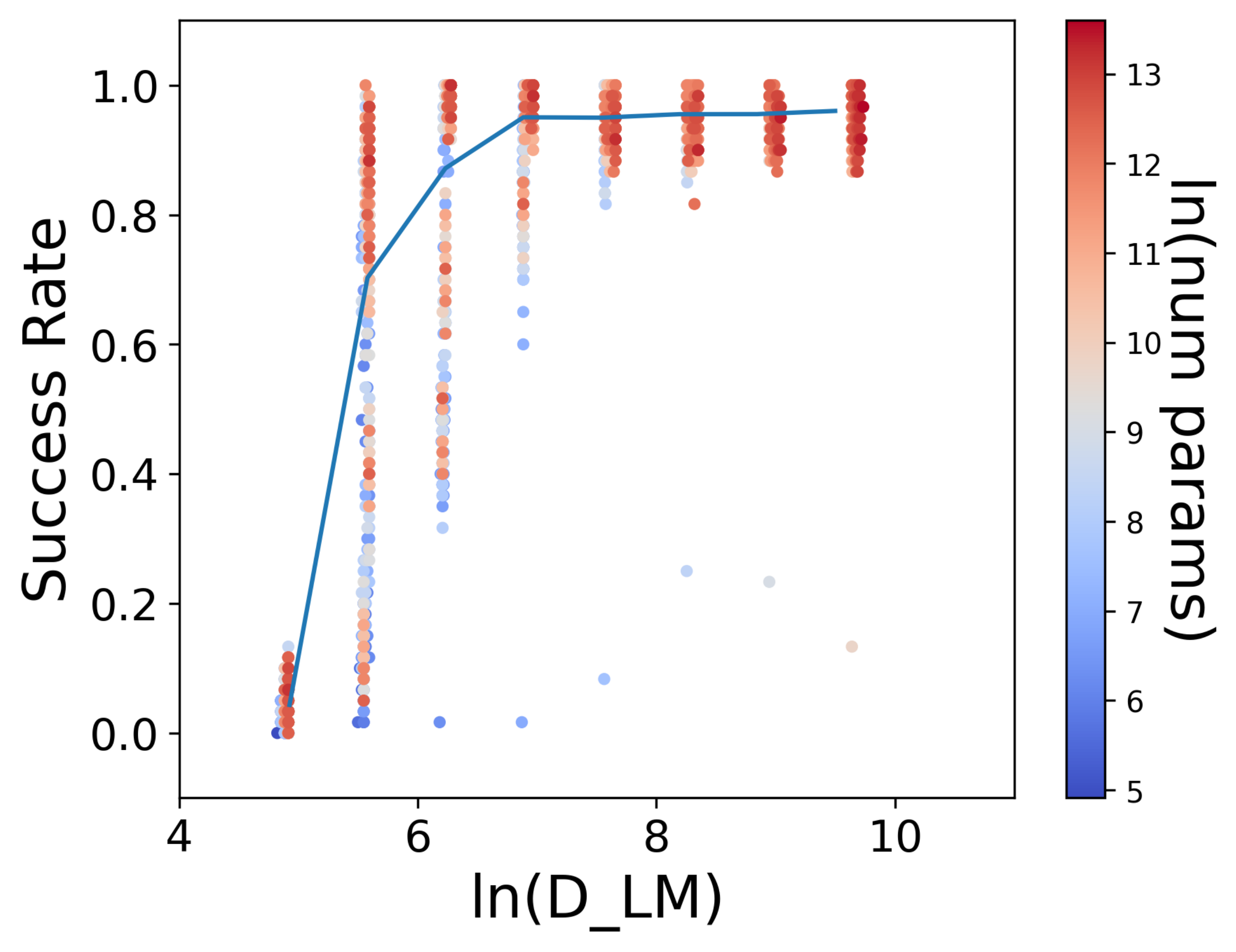}  \\
    \multicolumn{2}{c}{ (b) FetchPickAndPlace-v2}

\end{tabular}
\caption{\textbf{Performance correlated with capacity.} Each point represents a unique architecture, whose parameters are estimated by the hyper network, evaluated over multiple seeds. Left: Reward/Success rate vs $ln($number of paramters$)$; Right: Reward/Success rate vs $ln(D_{LM}$)}
 \label{fig:yes_cap}
 
\end{figure}

Table \ref{tab:perf} shows the performance of 90\% architectures. 
For comparison, we train MLPs with these architectures using off-policy RL. 
The performance of the 90\% architecture MLPs estimated by the GHP is on par with models trained during these runs, but finding these small architectures before training the GHP is nontrivial. We tried using behavior cloning to distill knowledge from the baseline [256,256] networks  to these architectures but the resulting performance was consistently poor.
We also tried using offline RL to distill polices into smaller networks, but after repeated hyperparameter searches did not obtain good results.

\subsection{Optimal Architecture Selection}
Can we predict an optimal architecture with a constraint on the total number of parameters? We analyze the correlation between reward and number of parameters in the architecture. After a complete training run on the Hopper task, we evaluate each architecture trained with the GHP.
Figure~\ref{fig:yes_cap}(a), shows that there is no significant correlation between the number of parameters and eventual performance. We compute $D_{LM}$ for each architecture (Algorithm~\ref{alg:cap}). Figure~\ref{fig:yes_cap}(b) shows a strong cut off in the $D_{LM}$ values, beyond which we get peak performance. We see such correlation in all the tasks. 
Inspired by this, we perform optimal architecture prediction for a given parameter constraint, by using the method described in Section~\ref{sec:method} - \ie we predict that the best architecture given a constraint on the number of parameters, maximizes $D_{LM}$. To test this hypothesis we sample $5000$ random cutoff values between $[ 1000, 1000000]$. 
Then, among all architectures with number of parameters less than these cutoff values, we predict the optimal architecture using Equation~\ref{eq:best_mlp}. We track its normalized performance relative to the actual best performing architecture, and average it for every sampled cutoff value.
As a baseline, we also compute the performance of the largest architecture that satisfies these cutoff constraints. Table~\ref{tab:best_mlp} shows that in almost all tasks, maximizing $D_{LM}$ produces an architecture whose performance is close to the optimal architecture that satisfies parameter count constraints, and is better than a naively chosen largest network.

\begin{table}[h]
        \centering
        \begin{tabular}{l | l | l }
        Task & Max Size & Max $D_{LM}$ \\
        \hline
        Ant-v2 & 0.73 & \textbf{0.78} \\
        HalfCheetah-v2 & 0.96  & \textbf{0.98} \\
        Walker2d-v2 & \textbf{0.90} & \textbf{0.90} \\
        Hopper-v2 & 0.98 & \textbf{0.99} \\
        Humanoid-v2 & 0.89 & \textbf{0.95} \\
        
        FetchReach-v1 & 0.99 & \textbf{1.0} \\
        FetchSlide-v1 & 0.75 & \textbf{0.80} \\
        FetchPush-v1 & 0.75 & \textbf{0.78} \\
        FetchPickAndPlace-v1 & 0.93 & \textbf{0.95} \\
        \end{tabular}
       \caption{\textbf{Architecture prediction.} For a given constraint on the number of parameters, ratio of performance of architectures (that maximize the lossless memory dimension; that maximize size) to that of best performing architecture. Closer to 1 is better.}
       \vspace{-0.15in}
       \label{tab:best_mlp}
\end{table}

Finally, we note some limitations in our method. The GHN used to model a GHP is large. While our objective is to achieve smaller MLPs during execution time, the size of the actor network is much larger during training, increasing training time significantly. 
With our single GPU implementations, even though the sample complexity did not increase, the total training time increased by $\sim$5x.

\section{Conclusion and Future Work}
\label{sec:con}

We use graph hyper networks to learn graph hyper policies trained with off-policy reinforcement learning. We generate networks that are 100x smaller than commonly used networks yet encode policies comparable to those encoded by much larger networks trained on the same task. Our results across locomotion and manipulation tasks confirm that our method can be appended to any off-policy deep learning algorithm, without any change in hyperparameters. Further, we obtain an array of working policies, with differing numbers of parameters, allowing us to pick an optimal network for the memory constraints of a system. Training multiple policies is as sample efficient as training a single policy. Finally, we provide a method to select the best architecture, given a constraint on the number of parameters.
In future work, we plan to move to image based observation space and move away from uniform sampling of architectures. %

\newpage
\bibliographystyle{IEEEtran}
\bibliography{refs}

\begin{thebibliography}{10}
\providecommand{\url}[1]{#1}
\csname url@samestyle\endcsname
\providecommand{\newblock}{\relax}
\providecommand{\bibinfo}[2]{#2}
\providecommand{\BIBentrySTDinterwordspacing}{\spaceskip=0pt\relax}
\providecommand{\BIBentryALTinterwordstretchfactor}{4}
\providecommand{\BIBentryALTinterwordspacing}{\spaceskip=\fontdimen2\font plus
\BIBentryALTinterwordstretchfactor\fontdimen3\font minus
  \fontdimen4\font\relax}
\providecommand{\BIBforeignlanguage}[2]{{%
\expandafter\ifx\csname l@#1\endcsname\relax
\typeout{** WARNING: IEEEtran.bst: No hyphenation pattern has been}%
\typeout{** loaded for the language `#1'. Using the pattern for}%
\typeout{** the default language instead.}%
\else
\language=\csname l@#1\endcsname
\fi
#2}}
\providecommand{\BIBdecl}{\relax}
\BIBdecl

\bibitem{haarnoja2019learning}
T.~Haarnoja, S.~Ha, A.~Zhou, J.~Tan, G.~Tucker, and S.~Levine, ``Learning to
  walk via deep reinforcement learning.'' in \emph{Robotics: Science and
  Systems}, 2019.

\bibitem{gu2017deep}
S.~Gu, E.~Holly, T.~Lillicrap, and S.~Levine, ``Deep reinforcement learning for
  robotic manipulation with asynchronous off-policy updates,'' in \emph{2017
  IEEE international conference on robotics and automation (ICRA)}.\hskip 1em
  plus 0.5em minus 0.4em\relax IEEE, 2017, pp. 3389--3396.

\bibitem{song2021autonomous}
Y.~Song, M.~Steinweg, E.~Kaufmann, and D.~Scaramuzza, ``Autonomous drone racing
  with deep reinforcement learning,'' in \emph{2021 IEEE/RSJ International
  Conference on Intelligent Robots and Systems (IROS)}.\hskip 1em plus 0.5em
  minus 0.4em\relax IEEE, 2021, pp. 1205--1212.

\bibitem{2017-TOG-deepLoco}
X.~B. Peng, G.~Berseth, K.~Yin, and M.~van~de Panne, ``Deeploco: Dynamic
  locomotion skills using hierarchical deep reinforcement learning,'' \emph{ACM
  Transactions on Graphics (Proc. SIGGRAPH 2017)}, vol.~36, no.~4, 2017.

\bibitem{batra2022decentralized}
S.~Batra, Z.~Huang, A.~Petrenko, T.~Kumar, A.~Molchanov, and G.~S. Sukhatme,
  ``Decentralized control of quadrotor swarms with end-to-end deep
  reinforcement learning,'' in \emph{Conference on Robot Learning}.\hskip 1em
  plus 0.5em minus 0.4em\relax PMLR, 2022, pp. 576--586.

\bibitem{ibarz2021train}
J.~Ibarz, J.~Tan, C.~Finn, M.~Kalakrishnan, P.~Pastor, and S.~Levine, ``How to
  train your robot with deep reinforcement learning: lessons we have learned,''
  \emph{The International Journal of Robotics Research}, vol.~40, no. 4-5, pp.
  698--721, 2021.

\bibitem{haarnoja2018soft}
T.~Haarnoja, A.~Zhou, P.~Abbeel, and S.~Levine, ``Soft actor-critic: Off-policy
  maximum entropy deep reinforcement learning with a stochastic actor,'' in
  \emph{International conference on machine learning}.\hskip 1em plus 0.5em
  minus 0.4em\relax PMLR, 2018, pp. 1861--1870.

\bibitem{DBLP:journals/corr/BrockmanCPSSTZ16}
\BIBentryALTinterwordspacing
G.~Brockman, V.~Cheung, L.~Pettersson, J.~Schneider, J.~Schulman, J.~Tang, and
  W.~Zaremba, ``Openai gym,'' \emph{CoRR}, vol. abs/1606.01540, 2016. [Online].
  Available: \url{http://arxiv.org/abs/1606.01540}
\BIBentrySTDinterwordspacing

\bibitem{JMLR:v20:18-598}
\BIBentryALTinterwordspacing
T.~Elsken, J.~H. Metzen, and F.~Hutter, ``Neural architecture search: A
  survey,'' \emph{Journal of Machine Learning Research}, vol.~20, no.~55, pp.
  1--21, 2019. [Online]. Available:
  \url{http://jmlr.org/papers/v20/18-598.html}
\BIBentrySTDinterwordspacing

\bibitem{pmlr-v100-mazoure20a}
\BIBentryALTinterwordspacing
B.~Mazoure, T.~Doan, A.~Durand, J.~Pineau, and R.~D. Hjelm, ``Leveraging
  exploration in off-policy algorithms via normalizing flows,'' in
  \emph{Proceedings of the Conference on Robot Learning}, ser. Proceedings of
  Machine Learning Research, L.~P. Kaelbling, D.~Kragic, and K.~Sugiura, Eds.,
  vol. 100.\hskip 1em plus 0.5em minus 0.4em\relax PMLR, 30 Oct--01 Nov 2020,
  pp. 430--444. [Online]. Available:
  \url{https://proceedings.mlr.press/v100/mazoure20a.html}
\BIBentrySTDinterwordspacing

\bibitem{stanley2009hypercube}
K.~O. Stanley, D.~B. D'Ambrosio, and J.~Gauci, ``A hypercube-based encoding for
  evolving large-scale neural networks,'' \emph{Artificial life}, vol.~15,
  no.~2, pp. 185--212, 2009.

\bibitem{DBLP:conf/iclr/HaDL17}
\BIBentryALTinterwordspacing
D.~Ha, A.~M. Dai, and Q.~V. Le, ``Hypernetworks,'' in \emph{5th International
  Conference on Learning Representations, {ICLR} 2017, Toulon, France, April
  24-26, 2017, Conference Track Proceedings}.\hskip 1em plus 0.5em minus
  0.4em\relax OpenReview.net, 2017. [Online]. Available:
  \url{https://openreview.net/forum?id=rkpACe1lx}
\BIBentrySTDinterwordspacing

\bibitem{zhang2018graph}
C.~Zhang, M.~Ren, and R.~Urtasun, ``Graph hypernetworks for neural architecture
  search,'' in \emph{International Conference on Learning Representations},
  2018.

\bibitem{knyazev2021parameter}
B.~Knyazev, M.~Drozdzal, G.~W. Taylor, and A.~Romero, ``Parameter prediction
  for unseen deep architectures,'' in \emph{Advances in Neural Information
  Processing Systems}, 2021.

\bibitem{todorov2012mujoco}
E.~Todorov, T.~Erez, and Y.~Tassa, ``Mujoco: A physics engine for model-based
  control,'' in \emph{2012 IEEE/RSJ International Conference on Intelligent
  Robots and Systems}.\hskip 1em plus 0.5em minus 0.4em\relax IEEE, 2012, pp.
  5026--5033.

\bibitem{DBLP:journals/corr/abs-1708-06019}
\BIBentryALTinterwordspacing
G.~Friedland and M.~M. Krell, ``A capacity scaling law for artificial neural
  networks,'' \emph{CoRR}, vol. abs/1708.06019, 2017. [Online]. Available:
  \url{http://arxiv.org/abs/1708.06019}
\BIBentrySTDinterwordspacing

\bibitem{sutton2018reinforcement}
R.~S. Sutton and A.~G. Barto, \emph{Reinforcement learning: An
  introduction}.\hskip 1em plus 0.5em minus 0.4em\relax MIT press, 2018.

\bibitem{lillicrap2016continuous}
T.~P. Lillicrap, J.~J. Hunt, A.~Pritzel, N.~Heess, T.~Erez, Y.~Tassa,
  D.~Silver, and D.~Wierstra, ``Continuous control with deep reinforcement
  learning.'' in \emph{ICLR (Poster)}, 2016.

\bibitem{andrychowicz2017hindsight}
M.~Andrychowicz, F.~Wolski, A.~Ray, J.~Schneider, R.~Fong, P.~Welinder,
  B.~McGrew, J.~Tobin, O.~Pieter~Abbeel, and W.~Zaremba, ``Hindsight experience
  replay,'' \emph{Advances in neural information processing systems}, vol.~30,
  2017.

\bibitem{DBLP:journals/corr/abs-1802-09464}
\BIBentryALTinterwordspacing
M.~Plappert, M.~Andrychowicz, A.~Ray, B.~McGrew, B.~Baker, G.~Powell,
  J.~Schneider, J.~Tobin, M.~Chociej, P.~Welinder, V.~Kumar, and W.~Zaremba,
  ``Multi-goal reinforcement learning: Challenging robotics environments and
  request for research,'' \emph{CoRR}, vol. abs/1802.09464, 2018. [Online].
  Available: \url{http://arxiv.org/abs/1802.09464}
\BIBentrySTDinterwordspacing

\end{thebibliography}

\end{document}